\documentclass[runningheads]{llncs}

\usepackage{eccv}
\usepackage{eccvabbrv}

\usepackage{graphicx}
\usepackage{booktabs}
\usepackage{array}
\usepackage{tikz}
\usetikzlibrary{arrows.meta,positioning,fit,backgrounds,calc,shapes.geometric}

\definecolor{acc}{RGB}{14,111,115}
\definecolor{soft}{RGB}{224,236,236}
\definecolor{drop}{RGB}{165,56,42}
\definecolor{gr}{RGB}{120,132,135}

\tikzset{
  bx/.style={draw=black!70, rounded corners=1.5pt, align=center,
             font=\scriptsize, inner sep=3.2pt, minimum height=7.5mm},
  io/.style={bx, fill=black!4},
  core/.style={bx, fill=soft, draw=acc, thick},
  gone/.style={bx, fill=white, draw=gr, dashed, text=gr},
  fl/.style={-{Stealth[length=1.6mm]}, draw=black!65},
  flacc/.style={-{Stealth[length=1.8mm]}, draw=acc, very thick},
  lbl/.style={font=\scriptsize\itshape, text=black!65},
  tag/.style={font=\tiny, text=gr},
  band/.style={rounded corners=2pt, draw=black!15},
}

\usepackage{hyperref}

\usepackage{fontawesome5}
\definecolor{linkblue}{RGB}{37,99,235}
\hypersetup{colorlinks=true,urlcolor=linkblue,linkcolor=black,citecolor=black}
\newcommand{\hflogo}{\raisebox{-0.28ex}{\includegraphics[height=1.0em]{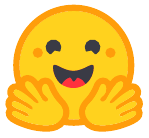}}}
\newcommand{\reslink}[4]{{#1~\textbf{#2:}~\href{#3}{#4}}}

\begin{document}

\title{Ambient @ EgoLongQA 2026: Distilling Long-Video perception into a Sub-2B Model}

\titlerunning{Ambient @ EgoLongQA 2026}

\author{Logesh Kumar Umapathi}
\authorrunning{L.\,K.\ Umapathi}
\institute{Team \textsc{Ambient} \\
\email{logeshkumaru@gmail.com}}

\maketitle

\begin{center}\vspace{-1mm}\small
\reslink{\faGithub}{GitHub}{https://github.com/ambient-intelligence-hq/egolongqa-2b}{github.com/ambient-intelligence-hq/egolongqa-2b}\\[2pt]
\reslink{\hflogo}{Hugging Face}{https://huggingface.co/collections/ambient-intelligence-labs/wearables-ai-workshop-eccv-2026}{Models \& Datasets Collection}
\end{center}

\begin{abstract}
We describe our entry to the EgoLongQA track of the Wearable-AI Challenge in ECCV 2026,
which placed \textbf{first in the ${\leq}2$B parameter division with 0.8279} on
the held-out test set. Our system is a single 2B vision-language model that
answers multiple-choice questions about ten-minute egocentric videos in one
greedy forward pass; It is obtained by distilling the
\emph{junior perception module} of a tool-using agentic pipeline, not the
agent itself into a small student, using teacher traces filtered to those that
answered correctly. it reaches 89\% of the accuracy of the large agentic pipeline using 1.1\% of its parameters. This raises a 27.1\% base model to 81.4\% on our held-out
questions. The 2B backbone has 2.2132\,B parameters and therefore over the
divisional limit, to make the entry admissable we prune the multilingual embedding table from 248{,}320 to
143{,}469 rows, reaching 1.9985\,B with provably identical logits on retained rows.
\end{abstract}

\section{Introduction}

The EgoLongQA task~\cite{wearableaiworkshop2026} asks a model to answer four-way
multiple-choice questions about long egocentric videos, typically around ten
minutes. Questions are often two-hop and compound: they refer to an anchor
event and then ask about something that happened relative to it. The ${\leq}2$B division constrains the entire
multimodal checkpoint, vision tower included.

A strong solution to this task could be an agent: sample frames, describe the video,
retrieve candidate clips with tools, and reason over the retrieved evidence. Our
own agentic pipeline, \textsc{Ambient}~\cite{ambient2026} --- three sampled
junior passes plus a senior orchestrator with clip-retrieval tools --- reaches
81.4\% on our distribution-robust evaluation set. This level of agentic pipeline performance is also entirely infeasible at 2B parameters

Our entry is the observation that the agent's accuracy can be moved into a small
single-pass student, provided one distils the \emph{right component}. Our solution involves three parts:

\begin{enumerate}
\item \textbf{Component-level distillation.} We distil~\cite{hinton2015distilling} the agent's
      junior perception module rather than the agent. Its input protocol is exactly
      what the 2B student receives at inference, giving zero train/test
      mismatch, whereas the senior's multi-turn tool trajectory is not
      expressible in a single forward pass (Sec.~\ref{sec:distill}).
\item \textbf{Admissibility by vocabulary pruning.} The embedding table is 23\%
      of the backbone. Removing unused rows takes the checkpoint from 2.2132\,B
      to 1.9985\,B with zero logit change on retained rows, and we document an
      asymmetry that makes this harder for generative models than for
      classifiers (Sec.~\ref{sec:prune}).
\item \textbf{A prior-corrected evaluation.} Language models are known to carry
      a selection bias over option identifiers~\cite{zheng2024mcqbias}; we observe
      that supervised distillation \emph{transfers the benchmark's own answer
      prior into the student}. We
      mitigate this by constructing our training and held-out sets with options permuted to a uniform gold distribution.
\end{enumerate}

\section{Method}

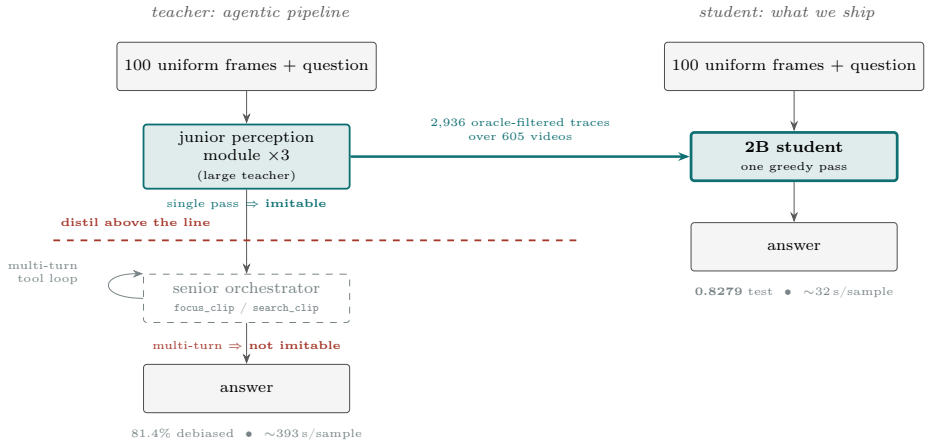
\begin{figure}[t]
\centering
\resizebox{\textwidth}{!}{%
\begin{tikzpicture}[x=1mm, y=1mm, font=\scriptsize]

\node[lbl, anchor=south west] at (-16,6) {teacher: agentic pipeline};
\node[io,   minimum width=32mm] (tin) at (0,0)    {100 uniform frames $+$ question};
\node[core, minimum width=32mm] (tj)  at (0,-14)  {junior perception\\ module $\times 3$\\ \tiny(large teacher)};
\node[gone, minimum width=32mm] (ts)  at (0,-36)  {senior orchestrator\\ \tiny\texttt{focus\_clip} / \texttt{search\_clip}};
\node[io,   minimum width=32mm] (ta)  at (0,-50)  {answer};

\draw[fl] (tin)--(tj);
\draw[fl] (tj)--(ts);
\draw[fl] (ts)--(ta);
\draw[fl, gr] (ts.west) .. controls +(-7,0) and +(-7,0) .. (ts.north west);
\node[tag, anchor=east, align=right] at (-25,-32) {multi-turn\\ tool loop};

\node[anchor=north, font=\tiny, text=acc]  at (0,-19.5) {single pass $\Rightarrow$ \textbf{imitable}};
\node[anchor=north, font=\tiny, text=drop] at (0,-41.5) {multi-turn $\Rightarrow$ \textbf{not imitable}};
\node[tag, anchor=north] at (0,-55) {81.4\% debiased\ \ $\bullet$\ \ $\sim$393\,s/sample};

\draw[dashed, drop, thick] (-30,-27) -- (52,-27);
\node[font=\tiny\bfseries, text=drop, anchor=south west] at (-30,-26) {distil above the line};

\node[lbl, anchor=south west] at (69,6) {student: what we ship};
\node[io,   minimum width=32mm] (sin) at (85,0)    {100 uniform frames $+$ question};
\node[core, minimum width=32mm, very thick] (sj) at (85,-14) {\textbf{2B student}\\ \tiny one greedy pass};
\node[io,   minimum width=32mm] (sa)  at (85,-28)  {answer};
\draw[fl] (sin)--(sj);
\draw[fl] (sj)--(sa);
\node[tag, anchor=north, align=center] at (85,-33) {\textbf{0.8279} test\ \ $\bullet$\ \ $\sim$32\,s/sample};

\draw[flacc, line width=1.1pt] (tj.east) -- (sj.west);
\node[font=\tiny, text=acc, align=center, anchor=south] at (42.5,-12.5)
      {2{,}936 oracle-filtered traces\\ over 605 videos};

\end{tikzpicture}}
\caption{The design decision. The agentic teacher pipeline has two components, the junior perception module consumes
frames and emits an answer in one shot. It is sampled 3 times, 
whenever there is no consensus between the three junior passes, the senior is called to resolve the ambiguity. 
The senior's value \emph{is} its multi-turn tool loop. We therefore distil the junior and discard the senior
rather than imitating it. Because the junior's input/output contract is exactly
what the student is served at inference and
the $\sim$393\,s orchestration collapses into one $\sim$32\,s generation.}
\label{fig:approach}
\end{figure}

\subsection{The agentic teacher pipeline}
\label{sec:teacher}

Fig.~\ref{fig:approach} shows the overall design.The teacher pipeline has two components. A junior perception module
consumes sampled frames with the question and options, and emits reasoning and an answer. It is sampled 3 times, 
if there is consensus between all the three junior passes, the answer is accepted and submitted. If there are no consensus, the senior -- a bigger model with tool calling
capabilities is called to resolve the ambiguity. The senior has access to tools to percieve intervals of the video in high fidelity and can use test time scaling to resolve the ambiguity and submit the final answer.

This agentic pipeline is our submission to the Large model division of the EgoLongQA track. We distill the traces of the junior perception module into a 2B student.

\subsection{The Student distillation}
\label{sec:contract}

The student is a 2B vision--language model~\cite{qwen35} from the Qwen3.5
family. We distill the perception module of the teacher agentic pipeline into the 2B student (along with additional synthetic data).

Qwen3.5 model's native dynamic-resolution vision tower and multimodal rotary
position encoding~\cite{wang2024qwen2vl} let the same adapter be
served at a higher resolution than it was trained at (Sec.~\ref{sec:distill}).

It consumes 100 uniformly sampled frames spanning the whole video, scaled to a
maximum dimension of 768, together with the question and its four options. It emits a timestamped free-text description followed by a structured
answer:

\begin{center}
\small
\texttt{<video\_description>}\,\ldots\,\texttt{</video\_description>}\\
\texttt{<answer><choice>X</choice><reason>\ldots</reason>}\\
\texttt{\phantom{<answer>}<citation>\ldots</citation></answer>}
\end{center}

This is verbatim the tool contract of the agent's preempt-answer
tool~\cite{ambient2026}.

\subsection{Oracle-filtered trace distillation}
\label{sec:distill}

Teacher models execute the junior contract over the training videos as the
perceiving step of the agent~\cite{ambient2026}, and the resulting traces are
harvested from its recorded trajectories. We retain only those passes whose
emitted \texttt{<choice>} matches ground truth and all three sampled junior passes agree on the answer. This
yields 2{,}936 traces over 605 videos, with a median of five distinct
descriptions per question. All teachers are open-weight
(Table~\ref{tab:teachers}); no closed model was used anywhere in the pipeline, at
a measured cost of about $1.4$ points against the best closed junior we
evaluated.

\begin{table}[t]
\centering
\caption{Teacher composition of the distillation set, every teacher is
open-weight.
$100$\,f\,/\,$400$\,f is that run's frame budget. Counts are \emph{post}-filter,
i.e.\ only passes whose \texttt{<choice>} matched gold, so they are lower than
the number of passes executed.}
\label{tab:teachers}
\setlength{\tabcolsep}{6pt}\small
\begin{tabular}{llrr}
\toprule
Teacher & Configuration & Traces & Share \\
\midrule
\texttt{google/gemma-4-31b-it}      & 630 videos, 100\,f & 1{,}233 & 42.0\% \\
\texttt{qwen/qwen3.5-122b-a10b}     & 700 videos, 100\,f & 1{,}178 & 40.1\% \\
\texttt{Qwen/Qwen3.6-27B}           & 700 videos  & 254 & 8.7\% \\
\texttt{google/gemma-4-31b-it}      & 70-video subset    & 145 & 4.9\% \\
\texttt{Qwen/Qwen3.5-122B-A10B} & 400\,f      & 76 & 2.6\% \\
\texttt{Qwen/Qwen3.6-35B-A3B}   & 400\,f      & 28 & 1.0\% \\
\texttt{Qwen/Qwen3.6-35B-A3B}   & 100\,f      & 22 & 0.7\% \\
\midrule
\multicolumn{2}{l}{Total} & \textbf{2{,}936} & 100\% \\
\bottomrule
\end{tabular}
\end{table}

\paragraph{The motivation for our distillation strategy.} Several teacher
runs were executed under a \emph{gated cascade}: three junior passes are sampled,
and if all three agree the pipeline accepts that answer and \textbf{skips the
senior entirely}; only non-unanimous samples are escalated to the senior with its
clip tools. Measuring the two paths separately is what pointed us at the junior
in the first place:

\begin{center}
\small
\begin{tabular}{lrr}
\toprule
Path & Share of samples & Accuracy \\
\midrule
junior unanimous (senior skipped) & 60--66\% & \textbf{91--95\%} \\
escalated to senior $+$ tools     & 34--41\% & 63--67\% \\
\bottomrule
\end{tabular}
\end{center}

Two thirds of the validation set is already solved, at over $90\%$, by junior passes
alone --- the senior only ever sees the residual, and scores poorly on it because
that residual is precisely the hard subset. This is the empirical reason we distil
the junior rather than the pipeline: most of the accuracy comes from the junior passes.

\subsection{Synthetic data}
\label{sec:synth}

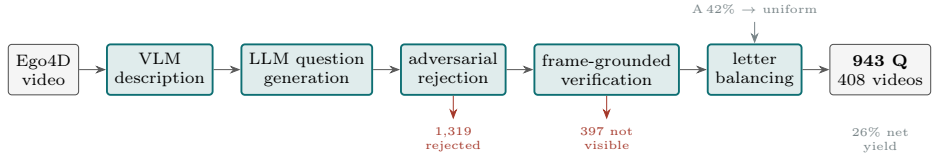
\begin{figure}[t]
      \centering
      \resizebox{\textwidth}{!}{%
      \begin{tikzpicture}[node distance=2.6mm]
      
      \node[io]   (vid)  {Ego4D\\ video};
      \node[core] (desc) [right=4mm of vid]  {VLM\\ description};
      \node[core] (gen)  [right=4mm of desc] {LLM question\\ generation};
      \node[core] (adv)  [right=4mm of gen]  {adversarial\\ rejection};
      \node[core] (ver)  [right=4mm of adv]  {frame-grounded\\ verification};
      \node[core] (bal)  [right=4mm of ver]  {letter\\ balancing};
      \node[io]   (out)  [right=4mm of bal]  {\textbf{943 Q}\\ 408 videos};
      
      \foreach \a/\b in {vid/desc, desc/gen, gen/adv, adv/ver, ver/bal, bal/out}
        \draw[fl] (\a)--(\b);
      
      \node[tag, text=drop, below=3.5mm of adv, align=center] (d1) {1{,}319\\rejected};
      \node[tag, text=drop, below=3.5mm of ver, align=center] (d2) {397 not\\visible};
      \draw[fl, drop] (adv.south)--(d1.north);
      \draw[fl, drop] (ver.south)--(d2.north);
      
      \node[tag, above=3.2mm of bal, align=center] (bn) {A\,42\%\ $\rightarrow$\ uniform};
      \draw[fl, gr] (bn.south)--(bal.north);
      
      \node[tag, below=3.5mm of out, align=center] {26\% net\\yield};
      
      \end{tikzpicture}}
      \caption{Synthetic MCQ generation. Two filters do the work: an adversarial pass
      that rejects questions answerable without the video, and a frame-grounded check
      that the evidence is actually visible. Net yield is 26\%. The balancing step is
      not cosmetic --- raw generator output is 42\% ``A'', which is trivially
      learnable.}
      \label{fig:synth}
      \end{figure}

Using the Ambient agent~\cite{ambient2026} we generated 943 audited
multiple-choice questions over 408 Ego4D~\cite{grauman2022ego4d} videos through
the pipeline of Fig.~\ref{fig:synth}, and used them to train the 2B student on
synthetic supervision alone. As shown in Table~\ref{tab:ood}, the synthetic set
\emph{teaches the task}: 27.1\% $\rightarrow$ 54.3\% on unseen videos. 
It has shown to help improve out-of-distribution robustness, and does
not absorb the answer prior (Sec.~\ref{paragraph:Prior corrected evaluation}).

\begin{table}[t]
\centering
\caption{Out-of-distribution behaviour on 500 human-written egocentric questions
with near-uniform gold~\cite{baermann2022qaego4d}. ``Predicted C-rate'' is how
often each model answers C, against a gold rate of 27\%: it measures whether the
validation set's answer prior was carried across. The trace-distilled model imports it (40\%); the synthetic-only
model does not (26\%).}
\label{tab:ood}
\setlength{\tabcolsep}{5pt}\small
\begin{tabular}{lrrr}
\toprule
Model & Accuracy & Predicted C-rate \\
\midrule
2B base                           & 49.4\% & 19\% \\
2B $+$ distillation (real traces) & 53.8\% & 40\% \\
\textbf{2B $+$ distillation + synthetic}    & \textbf{56.6\%} & \textbf{26\%} \\
\bottomrule
\end{tabular}
\end{table}

\subsection{Training.}
\label{sec:training}
 Training is LoRA~\cite{hu2022lora} ($r{=}16$, $\alpha{=}32$) for one epoch with a cosine schedule,
batch size 1 with gradient accumulation 8, and a $3\times$ token weight on the
\texttt{<choice>} span. Dropping that weight to $1.0$ was worse on every
evaluation we ran; 

\paragraph{Prior corrected evaluation.}
\label{paragraph:Prior corrected evaluation}
 The validation gold distribution is severely skewed,
Always answering C scores 63.4\%. We therefore constructed a held-out set of videos and
the same questions, with options permuted to a uniform gold distribution ---
the same permutation device used by \cite{zheng2024mcqbias} to estimate
inference-time selection bias, applied here as an evaluation rather than a
correction. It isolates perception from prior on identical content. All our 
experiments are measured against this debiased held-out set.

\paragraph{Resolution transfer.} We train at \texttt{max\_pixels}${=}50176$
($\approx$64 vision tokens per frame) and infer at $331776$ ($\approx$423 tokens
per frame), which is worth $+5.8$ points over training and inferring at the low
setting. The adapter learns the task rather than the resolution, so training cost
falls roughly $6\times$ at no accuracy penalty.

\paragraph{One epoch.} Epoch~1 outperformed epoch~2 in all five distillation runs
we conducted. On the option-shuffled model the second epoch costs $4.3$ points on
the debiased metric while \emph{raising} the skewed one, because it re-absorbs
the prior that shuffling removed.

\subsection{Vocabulary pruning to meet the parameter limit}
\label{sec:prune}

The backbone is 2.2132\,B parameters, over the divisional limit. Its composition
is shown in Table~\ref{tab:params}. The embedding table alone is 23\%, because
the vocabulary is 248{,}320 rows wide for multilingual coverage the task does not
require, and weight tying means each removed row saves its parameters once.

\begin{table}[t]
\centering
\caption{Parameter budget of the 2B backbone. Weight tying means the embedding
table is counted once, so each pruned row is a direct saving.}
\label{tab:params}
\begin{tabular}{lrr}
\toprule
Component & Parameters & Share \\
\midrule
Language layers      & 1{,}373.3\,M & 62.0\% \\
Embedding table (248{,}320 $\times$ 2048) & 508.6\,M & 23.0\% \\
Vision tower         & 331.4\,M & 15.0\% \\
\midrule
Total                & \textbf{2.2132\,B} & over limit \\
After pruning        & \textbf{1.9985\,B} & admissible \\
\bottomrule
\end{tabular}
\end{table}

We keep token ids $[0, 143000)$ unchanged, append a small set of explicitly
retained rare tokens, and relocate the 33 added and special tokens, copying every
embedding row exactly. The transformer, vision tower, projector and tokenizer are
untouched; the tokenizer still emits ids in the original space and the model code
remaps them before the embedding lookup. The result is 143{,}469 rows,
1.9985\,B parameters, a margin of 1.50\,M under the limit, and a measured logit
difference of exactly $0$ on all retained rows. Generation is byte-identical on
70 of 70 held-out items at serving resolution.

\section{Results}

\begin{table}[t]
\centering
\caption{Main results. All figures use 70 held-out videos disjoint from every
training set, 100 frames, greedy decoding at \texttt{max\_pixels}${=}331776$.}
\label{tab:main}
\setlength{\tabcolsep}{8pt}
\begin{tabular}{llrr}
\toprule
Model & Training data & Held-out Accuracy \\
\midrule
Qwen3.5-2B base       & ---                            & 27.1\% \\
\;+ synthetic only    & 943 generated questions        & 54.3\% \\
\;+ distillation      & 2{,}936 oracle-filtered traces & \textbf{81.4\%} \\
\;+ distill\,+\,synthetic & 3{,}879 rows               & \textbf{81.4\%} \\
\midrule
27B distillation      & same 2{,}936 traces            & 85.7\% \\
27B base              & ---                            & 14.3\% \\
\bottomrule
\end{tabular}
\end{table}

Table~\ref{tab:main} summarises. Distillation is the dominant lever: it moves the
2B backbone from 27.1\% to 81.4\%, a $3\times$ improvement, and moves a 27B model
from a 65\% single-pass baseline to 85.7\% while replacing a three-vote,
gated, senior-agent pipeline with one greedy generation , reducing per-sample
latency from roughly 393\,s to 32\,s. Even though the synthetic data
did not help improve the held-out accuracy, it has shown to help improve out-of-distribution robustness (Table~\ref{tab:ood}).

\subsection{Final leaderboard}
\label{sec:leaderboard}

Table~\ref{tab:leaderboard} gives the official held-out test results for the
${\leq}2$B division, as published on the challenge
leaderboard~\cite{wearablesleaderboard2026}. Our entry placed first at
\textbf{0.8279}, a margin of $1.27$\,pp over the runner-up.

\begin{table}[t]
\centering
\caption{Official ${\leq}2$B leaderboard, EgoLongQA~\cite{wearablesleaderboard2026}.
All entries are open-weight. Accuracy is on the organisers' held-out test set.}
\label{tab:leaderboard}
\setlength{\tabcolsep}{6pt}\small
\begin{tabular}{rllrr}
\toprule
\# & Team & Model & Params & Accuracy \\
\midrule
\textbf{1} & \textbf{ambient} & \textbf{ambient-agent-small-distill-v1} &
\textbf{1.99\,B} & \textbf{0.8279} \\
2 & smartkaist   & Qwen-Atomic        & 1.99\,B & 0.8152 \\
3 & sololevelling& internvl1b         & 1.24\,B & 0.7804 \\
4 & outofmemory  & Test-Pipeline-V2   & 1.30\,B & 0.7730 \\
5 & hippo        & HIPPO-merge        & 1.30\,B & 0.7434 \\
6 & fufu         & EgoAssist-LongQA-0.8B-OptStacker & 0.85\,B & 0.3728 \\[-1pt]
7 & deuxvoir     & Vertex             & 1.87\,B & 0.2967 \\
\bottomrule
\end{tabular}
\end{table}

\section{Negative results}

We report these at length because they consumed most of the project and several
are, in our view, more useful than the positive results

\subsection{Reinforcement learning}

Our experiments with reinforcement learning with verifiable reward ( using GRPO)
~\cite{shao2024grpo} over our SFT distilled modeldid not yield any reproducible held-out gain. We designed the rewards 
such that if the rollouts selected the correct option (from shuffled options) 
then the reward was 1, otherwise it was 0.

We did not find the remedy in the algorithm. Subsequent work addresses several biases we might otherwise suspect: 
DAPO introduces decoupled clipping, dynamic sampling, and token-level loss aggregation~\cite{yu2025dapo},
while Dr.\ GRPO removes the response-length and group reward-variance normalizations responsible for length and difficulty biases~\cite{liu2025drgrpo}.
Neither, however, addresses our actual constraint.

We suspect that the reason for the lack of gain is lack of in-distribution and unseen prompts for rollouts. 
In-distribution prompts used for rolloutswere seen during supervised training, so reward gains there are memorisation. 
Out-of-distribution prompts do not transfer. Prompts that are both in-distribution and unseen number roughly 25
videos, which is insufficient.

\subsection{Frame budget: the sign of the effect depends on model capacity}
\label{sec:frames}

Showing the model more of the video is the most intuitive lever available, and it
is the one whose result surprised us most: \emph{more frames help a mid-size
model and hurt a small one}. We measured both directly.

\paragraph{At ${\leq}2$B, more frames hurt.} Comparing 100 and 400 frames on the
same question indices, with thinking enabled in both arms:

\begin{center}
\begin{tabular}{lr}
\toprule
Setting & Full-700 accuracy \\
\midrule
100 frames & \textbf{64.3\%} \\
400 frames & 61.5\% \quad ($-4.4$\,pp) \\
\bottomrule
\end{tabular}
\end{center}

\paragraph{At 35B, the same change helps.} We observed a +11.2 pp improvement in accuracy when using 400 frames instead of 100 frames with
Qwen3.6-35B-A3B model in our agentic pipeline for large division submission.

\section{Conclusion}

A 2B model can be brought to competitive long-form egocentric video QA by
distilling the perception component of an agentic pipeline rather than the agent,
and can be made admissible under a strict parameter limit by pruning vocabulary
the task does not use.

The entry placed first, and the test score matched our
validation figure to within $1.4$\,pp.Further, The robustness to out-of-distribution 
videos is achieved by constructing a synthetic dataset from Ego4D and 
distilling it into the student.

\section*{Acknowledgements}
We thank the challenge organisers for the benchmark and the evaluation
infrastructure. The authors declare no competing interests.

\bibliographystyle{splncs04}
\bibliography{refs}

\end{document}